\documentclass{article}
\usepackage{ijcai25}

\usepackage{times}
\usepackage{soul}
\usepackage{url}
\usepackage[hidelinks]{hyperref}
\usepackage[utf8]{inputenc}
\usepackage[small]{caption}
\usepackage{graphicx}
\usepackage{amsmath}
\usepackage{amsthm}
\usepackage{booktabs}
\usepackage{algorithm}
\usepackage{algorithmic}
\usepackage[switch]{lineno}

\newcommand{\name}{RobustX}
\usepackage{newfloat}
\usepackage{color}
\usepackage{listings}

\definecolor{codegreen}{rgb}{0,0.6,0}
\definecolor{codegray}{rgb}{0.5,0.5,0.5}
\definecolor{codepurple}{rgb}{0.58,0,0.82}
\definecolor{backcolour}{rgb}{0.95,0.95,0.92}

\lstdefinestyle{mystyle}{
    backgroundcolor=\color{backcolour},   
    commentstyle=\color{codegreen},
    keywordstyle=\color{magenta},
    numberstyle=\tiny\color{codegray},
    stringstyle=\color{codepurple},
    basicstyle=\ttfamily\footnotesize,
    breakatwhitespace=false,         
    breaklines=true,                 
    captionpos=b,                    
    keepspaces=true,                 
    numbers=left,                    
    numbersep=5pt,                  
    showspaces=false,                
    showstringspaces=false,
    showtabs=false,                  
    tabsize=2
}

\usepackage{subcaption}

\usepackage{todonotes}

\title{\name: Robust Counterfactual Explanations Made Easy}

\author{
Junqi Jiang$^1$
\and
Luca Marzari$^2$\and
Aaryan Purohit$^1$\And
Francesco Leofante$^1$\\
\affiliations
$^1$ Department of Computing, Imperial College London, United Kingdom\\
$^2$University of Verona, Department of Computer Science, Italy \\
\emails
\{junqi.jiang20, francesco.leofante\}@imperial.ac.uk,
luca.marzari@univr.it
}

\begin{document}

\maketitle

\begin{abstract}
   The increasing use of Machine Learning (ML) models to aid decision-making in high-stakes industries demands explainability to facilitate trust. Counterfactual Explanations (CEs) are ideally suited for this, as they can offer insights into the predictions of an ML model by illustrating how changes in its input data may lead to different outcomes. 
   However, for CEs to realise their explanatory potential, significant challenges remain in ensuring their robustness under slight changes in the scenario being explained. Despite the widespread recognition of CEs' robustness as a fundamental requirement, a lack of standardised tools and benchmarks hinders a comprehensive and effective comparison of robust CE generation methods. In this paper, we introduce \name, an open-source Python library implementing a collection of CE generation and evaluation methods, with a focus on the robustness property. \name{} provides interfaces to several existing methods from the literature, enabling streamlined access to state-of-the-art techniques. The library is also easily extensible, allowing fast prototyping of novel robust CE generation and evaluation methods. 
\end{abstract}

\section{Introduction}
\label{sec:intro}

With the increasing use of Machine Learning (ML) models to aid decision-making in high-stakes fields such as healthcare~\cite{shaheen2021applications} and finance~\cite{cao2022ai}, there is a growing need for better explainability of these models. Counterfactual Explanations (CEs)~\cite{guidotti2024counterfactual} are often leveraged in Explainable AI (XAI) to this end due to their intelligibility and alignment with human reasoning~\cite{Miller_19,Byrne19}. In particular, CEs can offer insights into the predictions produced by an ML model by showing how small changes in its input may lead to different (often more desirable) outcomes. To see what benefits CEs can bring, consider an illustration of a loan application with features \emph{27} years of age, \emph{low} credit rating, and \emph{15K} loan amount. Assume a bank's ML model classifies the application as not creditworthy. A CE for this outcome could be an altered input where a \emph{medium} credit rating (with the other features unchanged) would result in the application being classified as creditworthy, thus giving the applicant an idea of what is required to have their loan approved.

\begin{figure}[t!]
    \begin{subfigure}[b]{0.23\textwidth} 
        \centering
        \fbox{\includegraphics[width=\textwidth, clip, trim={5cm 3.5cm 5cm 4.5cm}]{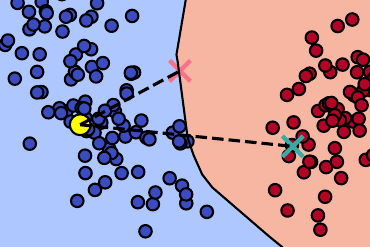}}
        \caption{}
        \label{fig:sub1} 
    \end{subfigure}
    \hspace{0.1cm}
    \begin{subfigure}[b]{0.23\textwidth} 
        \centering
        \fbox{\includegraphics[width=\textwidth, clip, trim={5cm 3.5cm 5cm 4.5cm}]{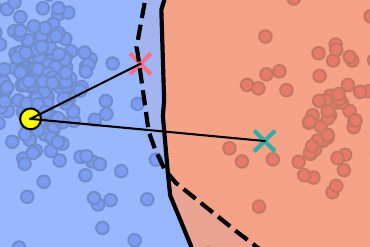}}
        \caption{}
        \label{fig:sub2} 
    \end{subfigure}
    \caption{A lack of robustness may invalidate CEs, here demonstrated on a neural network classifier trained to solve a binary classification task. An input (yellow circle) receives an initial classification, and two counterfactuals (red and green crosses) are generated for it (Figure~\ref{fig:sub1}). After a fine-tuning step occurs (Figure~\ref{fig:sub2}), the decision boundary slightly changes (from dashed to full black line), and previously generated CEs may be invalidated (red cross).}
    \label{fig:main} 
\end{figure}

Despite their potential, current approaches to generating CEs often fall short in generating \emph{robust explanations}. Consequently, these methods may produce explanations whose validity is compromised by slight changes in the scenario being explained. For instance, recent work~\cite{UpadhyayJL21,JiangL0T23,DBLP:conf/icml/HammanNMMD23} has highlighted that even small alterations in the parameters of an ML model, e.g. following fine-tuning, may invalidate previously generated CEs. An example of this scenario is captured in Figure~\ref{fig:main}, where a lack of robustness is demonstrated on a model trained for binary classification tasks. In Figure~\ref{fig:sub1}, an input (yellow circle) receives an initial classification (blue class), and two counterfactuals are generated for it: one (red cross) laying exactly on the decision boundary (full black line) and one deeper inside the counterfactual class (green cross). In Figure~\ref{fig:sub2}, we observe that the decision boundary of the model undergoes slight changes, induced by fine-tuning on a slightly shifted input distribution. As a result, previously generated CEs may cease to be valid if no precautions are taken to ensure robustness. For instance, we observe that the CE corresponding to the red cross is now classified as belonging to the blue class and is thus invalid. Now consider the consequences of these behaviours in our loan example: after fine-tuning, the applicant changing their credit rating to medium no longer ensures the success of the loan application, as the CE was not robust. When this happens, the CE previously generated by the bank is invalidated, and the bank may be liable for inconsistent statements made to customers regarding loan terms.

This and many other forms of robustness of CEs have been the subject of intense research efforts recently, and numerous algorithms to evaluate the robustness of CEs have been proposed (a recent survey identified about 40 methods~\cite{ijcai2024survey}). However, the current state of robust CE research is fragmented, with various methods developed independently and implemented in different, often incompatible, ways. This lack of standardisation has resulted in challenges for the broader research community, as comparing the effectiveness of robust CE generation methods is impractical. 
 
We fill this gap in this paper and introduce \name, an open-source Python library to standardise and streamline the generation, evaluation, and benchmarking of robust CEs. \name{} provides flexible, extensible, and customisable tools to implement custom CE methods. Differently from existing frameworks, e.g.~\cite{PawelczykBHRK21,AgarwalKSPJPZL22}, our library focuses on providing a consistent framework for testing robustness and systematically comparing various methods, ensuring fair and reliable evaluations. \name{} addresses key limitations on library tools in the current landscape (\cite{DBLP:conf/ijcai/KeaneKDS21,ijcai2024survey}) by offering a standardised approach to robust CE development while also promoting extensibility, allowing users to integrate new datasets and explanation algorithms as needed. The library, including documentation and tutorials, is publicly available at the following link: 
\begin{center}
    \textit{\url{https://github.com/RobustCounterfactualX/RobustX}}
\end{center}

The reminder of this paper is organised as follows. Section~\ref{sec:overview} presents the main components of the library, providing details about their functionalities. Section~\ref{sec:comparison} demonstrates how easy it is to use \name{} to benchmark existing CE generation algorithms and compare them using different metrics. Finally, Section~\ref{sec:conclusion} offers some concluding remarks and pointers for future work.

\section{Overview}
\label{sec:overview}
\begin{figure}[ht!]
    \centering
    \includegraphics[width=\linewidth, clip, trim={25.5cm 9.5cm 14.5cm 3.5cm}]{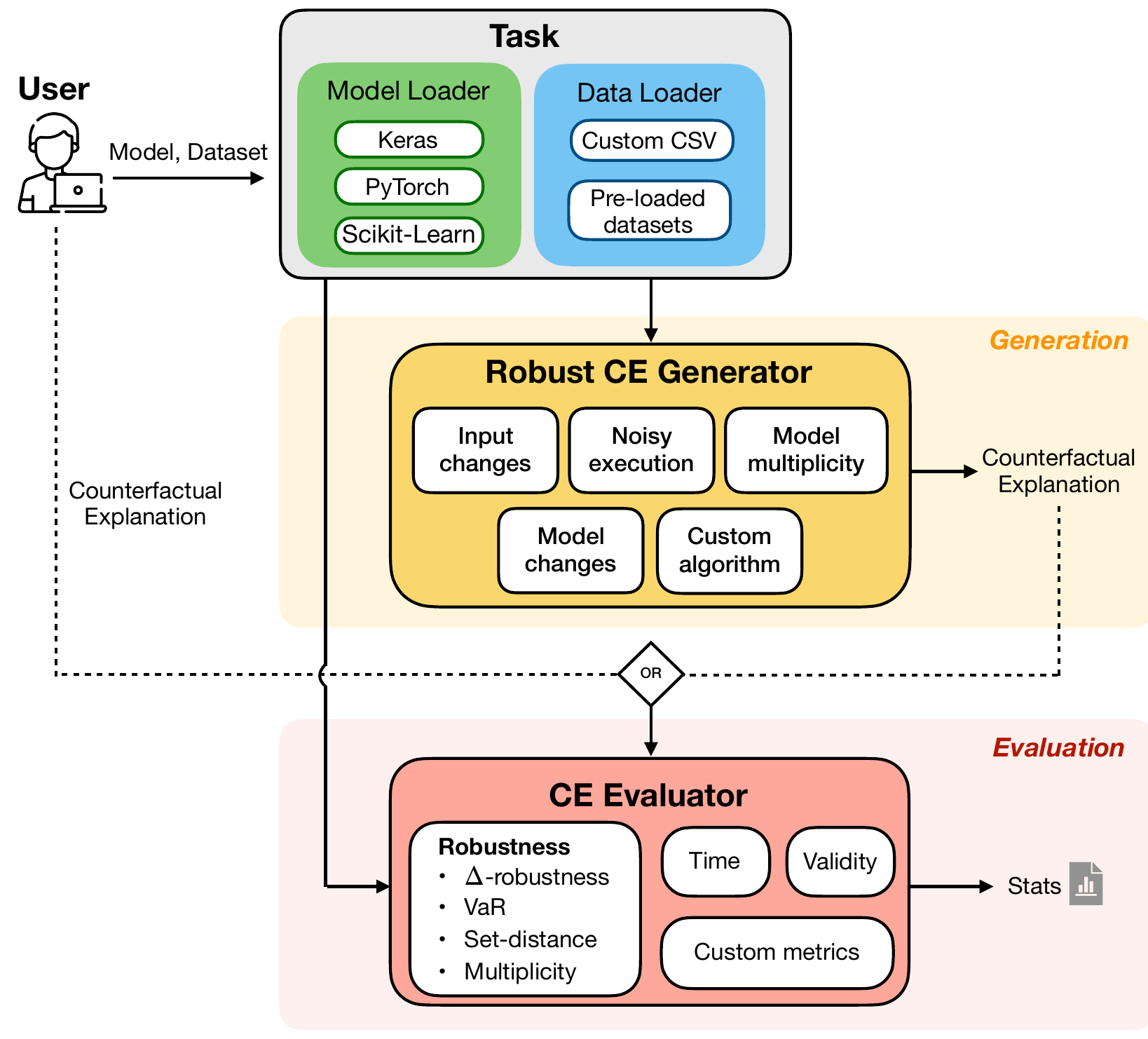}
    \caption{Internal work-flow of \name{}. Users can choose whether to generate robust CEs using our library or evaluate the robustness of externally generated CEs in our \name{}'s evaluation facilities. }
    \label{fig:robustx}
\end{figure}

RobustX implements a complete pipeline for robust CE generation and evaluation (Figure \ref{fig:robustx}) with three major components: \textbf{Task}, \textbf{CE generator}, and \textbf{CE evaluator}. Each has an abstract class template for easy customisation. Users start by creating a task which defines the model and inputs to be explained. Then, the user can choose whether to use \name{} to generate CEs, or directly evaluate the robustness of previously (externally) generated CEs.

\textbf{Task} objects, providing functionality for interactions between models and datasets, are the basic class passed into the CE generation and evaluation pipelines. Our current implementation assumes a \texttt{ClassificationTask} by default, as this is the most commonly considered use case in the literature; however, users can also implement customised \texttt{Task} objects for learning problems other than classification. \name{} natively supports models trained using sklearn\;\cite{pedregosa2011scikit}, Keras\;\cite{chollet2015keras} and PyTorch\;\cite{pytorch}. Models trained using other frameworks can also be used by instantiating a \texttt{BaseModel} wrapper class. As far as datasets are concerned, \name{} offers a selection of pre-loaded example datasets that can be readily loaded using the \texttt{DatasetLoader} class. Additionally, this class also allows the uploading of custom datasets if needed via \textit{.csv} files. 

\begin{figure*}[t!]
    \centering
    \begin{lstlisting}[language=Python]
# first prepare a task
from robustx.datasets.ExampleDatasets import get_example_dataset
from robustx.lib.models.pytorch_models.SimpleNNModel import SimpleNNModel
from robustx.lib.tasks.ClassificationTask import ClassificationTask

data = get_example_dataset("ionosphere")
data.default_preprocess()
model = SimpleNNModel(34, [8], 1)
model.train(data.X, data.y)
task = ClassificationTask(model, data)

# specify the names of the methods and evaluations we want to use, run benchmarking
# This will find CEs for all instances predicted with the undesirable class (0) and compare
from robustx.lib.DefaultBenchmark import default_benchmark

methods = ["KDTreeNNCE", "MCE", "MCER", "RNCE", "STCE", "PROPLACE"]
evaluations = ["Validity", "Distance", "Delta-robustness"]

default_benchmark(task, methods, evaluations, neg_value=0, column_name="target", delta=0.005)
\end{lstlisting}
    \caption{Code snippet exemplifying how \name{} can be used to easily benchmark CE methods. Table \ref{tab:results} lists the benchmarking results.
    }
    \label{fig:example}
\end{figure*}

\name{} currently implements nine \textbf{robust CE generation methods} across different robustness use cases reported in the yellow box in Figure\;\ref{fig:robustx}: \texttt{AP$\Delta$S} \cite{DBLP:conf/ecai/MarzariLCF24}, \texttt{ArgEnsembling} \cite{DBLP:conf/aamas/JiangL0T24}, \texttt{DiverseRobustCE} \cite{LeofanteP24}, \texttt{MCER} \cite{JiangL0T23}, \texttt{ModelMultiplicityMILP} \cite{DBLP:conf/kr/LeofanteBR23}, \texttt{PROPLACE} \cite{DBLP:conf/acml/JiangLL0T23}, \texttt{RNCE} \cite{JiangLRT24}, \texttt{ROAR} \cite{UpadhyayJL21}, \texttt{STCE} \cite{DBLP:conf/icml/DuttaLMTM22,DBLP:conf/icml/HammanNMMD23}. It also provides four popular non-robust methods that can be used as baselines to crease new generation methods: \texttt{BLS} \cite{LeofanteP24}, \texttt{MCE} \cite{MohammadiKBV21}, \texttt{KDTreeNNCE} \cite{DBLP:journals/datamine/BrughmansLM24}, and the seminal work by \cite{Wachter17}. All methods inherit from the abstract class \texttt{CEGenerator} and implement the  \texttt{\_generation\_method()} function, providing an easy interface to other components in the pipeline. 

\textbf{CE evaluation methods} can take in CEs, either generated within or outside \name{}, and benchmark their robustness along with other common properties identified in the literature. Currently, \name{} provides a \texttt{CEEvaluator} class to evaluate the validity and proximity of CEs~\cite{Wachter17}, as well as five classes specifically focusing on robustness evaluation metrics:
\texttt{VaRRobustnessEvaluator} to assess the validity of CEs after retraining~\cite{DBLP:conf/icml/DuttaLMTM22}, \texttt{DeltaRobustnessEvaluator}~\cite{JiangL0T23} to assess the robustness of CEs under plausible model changes, \texttt{ApproximateDeltaRobustnessEvaluator}~\cite{DBLP:conf/ecai/MarzariLCF24} assessing probabilistic robustness to plausible model changes, \texttt{SetDistanceRobustnessEvaluator} \cite{LeofanteP24} for assessing stability of CEs when the input is perturbed, and \texttt{MultiplicityValidityRobustnessEvaluator}~\cite{DBLP:conf/kr/LeofanteBR23} for checking CE robustness under model multiplicity. When needed, additional robustness evaluators can be easily added through the extensible interface provided by \name{}.
 

\section{\name{} in Action }
\label{sec:comparison}
In this section we provide an example on how to use \name{} in practice; additional examples are available online. Figure~\ref{fig:example} shows how to run and compare six methods supported by \name{}. In this example we focus on robustness against model changes~\cite{UpadhyayJL21} and perform a comparison between four robust methods and two non-robust baselines. To this end, we first import the (pre-loaded) \texttt{ionosphere} dataset for binary classification (line~6) and apply standard pre-processing. We then create and train a simple three-layer neural network model (line~8).
A task object is then created from the dataset and model. Then, we specify the CE generation and evaluation methods of interest (line~16 and 17) and run the benchmarking procedure (line~19). The default benchmark function in this example runs each method with its default hyperparameters, although customised hyperparameters can be configured. It then generates CEs for all instances in the dataset which are predicted with an undesirable class (here 102 points with neg\_value=0), and runs the specified evaluation methods. In this example, we evaluate CEs along three metrics: validity, proximity~\cite{Wachter17} and $\Delta$-robustness~\cite{JiangL0T23}. The results along the selected evaluation metrics are then printed in a structured table, so that 
we can easily compare how each method performs. 

Table \ref{tab:results} shows the results obtained, and reports the computation time, the percentage of valid CEs, average proximity (L2 distance to the input), and the percentage of CEs that are $\Delta$-robust. Observing this table, users of \name{} will be able to identify that robustness performance improvements from the non-robust baselines (NNCE, MCE) to the robust methods (MCER, RNCE, STCE, PROPLACE) are notable, although MCER fails to achieve 100\% robustness. Based on these results, users might conclude that \texttt{RNCE} is the optimal CE generator for this task, balancing between computation time, proximity, and robustness. 

\begin{table}[h]
\centering
\resizebox{\linewidth}{!}{
\begin{tabular}{l||cccc}
\hline
\textbf{Method}   & \textbf{Time (s)} & \textbf{Validity ($\%$)} & \textbf{Proximity} & \textbf{Rob. ($\%$)}\\\hline
KDTreeNNCE & 0.2 & 100 & 5.76 & 51.6 \\ 
MCE & 3.4 & 100 & 2.95 & 0 \\ 
MCER & 137.6 & 100 & 4.84 & 64.8 \\ 
RNCE & 3.9 & 100 & 6.03 & 100 \\ 
STCE & 39.7 & 100 & 7.30 & 100 \\ 
PROPLACE & 12.9 & 100 & 6.02 & 100 \\
\hline
\end{tabular}
}
\caption{Example benchmarking of six CE generation methods.}
\label{tab:results}
\vspace{-2mm}
\end{table}

\section{Conclusion}
\label{sec:conclusion}
We presented \name{}, a Python framework to generate, evaluate and compare robust CE for ML models. Our library fills a major gap in the existing literature on robust CEs, providing an easy-to-use and extensible platform to benchmark existing algorithms for robust CEs. Building upon extensive research in the area, \name{} provides a unified platform to run and compare existing approaches, as well as implementing new ones, reducing the need to re-implement software from scratch. Work is underway to further expand the list of available generation algorithms, evaluation methods and additional software facilities for testing and validation to ensure the correctness of the implementations. We believe \name{} will streamline research efforts in robust CEs, accelerating the development of innovative solutions and fostering collaboration within this rapidly growing field.

\newpage
\bibliographystyle{named}
\bibliography{references}

\end{document}